%% file: main.tex
\documentclass[conference]{IEEEtran}
\usepackage{cite}
\usepackage{amsmath,amssymb,amsfonts}
\usepackage{algorithmic}
\usepackage{graphicx}
\usepackage{textcomp}
\usepackage[table,xcdraw]{xcolor}
\usepackage{soul}
\usepackage{booktabs}
\usepackage{multirow}
\usepackage{hyperref}
\usepackage{tabularx}
\usepackage{color}
\usepackage{bbm}
\usepackage{svg}
\usepackage{ulem}
\usepackage{booktabs, siunitx, multirow}
\def\BibTeX{{\rm B\kern-.05em{\sc i\kern-.025em b}\kern-.08em
    T\kern-.1667em\lower.7ex\hbox{E}\kern-.125emX}}

\definecolor{lightgreen}{RGB}{184,222,218}
\definecolor{lightblue}{RGB}{202,220,236}
\newcommand{\greenhl}[1]{\sethlcolor{lightgreen}\hl{#1}}
\newcommand{\bluehl}[1]{\sethlcolor{lightblue}\hl{#1}}

\begin{document}
\title{Enhancing Reinforcement Learning for the Floorplanning of Analog ICs with Beam Search \\
\thanks{
This work has been developed in the project HoLoDEC (project label 16ME0696) which is partly funded within the Research Programme ICT 2020 by the German Federal Ministry of Education and Research (BMBF) and partially supported by the PNRR project iNEST (Interconnected North-Est Innovation Ecosystem) funded by the European Union Next-GenerationEU (Piano Nazionale di Ripresa e Resilienza (PNRR) – Missione 4 Componente 2, Investimento 1.5 – D.D. 1058 23/06/2022, ECS\_00000043).
}
}
\IEEEoverridecommandlockouts

\DeclareRobustCommand{\IEEEauthorrefmark}[1]{\smash{\textsuperscript{\footnotesize #1}}} 

\author{\IEEEauthorblockN{Sandro Junior Della Rovere\IEEEauthorrefmark{1,2},
Davide Basso\IEEEauthorrefmark{1,2}, Luca Bortolussi\IEEEauthorrefmark{1},
Mirjana Videnovic-Misic\IEEEauthorrefmark{2} and Husni Habal\IEEEauthorrefmark{3}}
\IEEEauthorrefmark{1}University of Trieste, Italy / \IEEEauthorrefmark{2}Infineon Technologies AT, Villach / \IEEEauthorrefmark{3}Infineon Technologies AG, Munich\\
sandrojunior.dellarovere@studenti.units.it, davide.basso@phd.units.it}

\maketitle\IEEEpubidadjcol
%
\hbadness=10000\maketitle\hbadness=1000

\makeatletter
\newcommand{\linebreakand}{%
  \end{@IEEEauthorhalign}
  \hfill\mbox{}\par
  \mbox{}\hfill\begin{@IEEEauthorhalign}
}
\makeatother
\bstctlcite{IEEEexample:BSTcontrol}
\maketitle

\begin{abstract}
\input{paper_body/abstract.tex}
\end{abstract}

\begin{IEEEkeywords}
Reinforcement Learning, Beam Search, Analog Circuits, Physical Design, Electronic Design Automation
\end{IEEEkeywords}

\section{Introduction}\label{introduction}
\input{paper_body/introduction}

\section{Problem Definition}\label{problem_definition}
\input{paper_body/problem_formulation}

\section{Beam Search Boosted RL Floorplanning}\label{bs_boosted_floorplanning}
\input{paper_body/methodology}

\section{Experimental Results and Discussion}\label{experimental_results_discussion}
\input{paper_body/results}

\section{Conclusions}\label{conclusions}
\input{paper_body/conclusions_future_research}

\bibliographystyle{IEEEtran} 
\bibliography{IEEEabrv,refs}

\end{document}

%% file: paper_body/abstract.tex
The layout of analog ICs requires making complex trade-offs, while addressing device physics and variability of the circuits.
This makes full automation with learning-based solutions hard to achieve.
However, reinforcement learning (RL) has recently reached significant results, particularly in solving the floorplanning problem.
This paper presents a hybrid method that combines RL with a beam search (BS) strategy.
The BS algorithm enhances the agent’s inference process, allowing for the generation of flexible floorplans by accommodating various objective weightings, and addressing congestion without the need for policy retraining or fine-tuning.
Moreover, the RL agent's generalization ability stays intact, along with its efficient handling of circuit features and constraints.
Experimental results show $\sim5-85\%$ improvement in area, dead space and half-perimeter wire length compared to a standard RL application, along with higher rewards for the agent.
Moreover, performance and efficiency align closely with those of existing state-of-the-art techniques.\looseness=-1

%% file: paper_body/introduction.tex

The physical design automation of analog integrated circuits (ICs) has progressed slowly in comparison to digital design.
Many iterations are needed for the layout engineer to achieve an optimal result where
strict topological constraints are satisfied,
device matching is achieved, and parasitic effects are minimized. 
Metaheuristics such as simulated annealing (SA) have been used in the past. However, they fail to leverage past information to intelligently navigate the solution space.
Recently, reinforcement learning (RL) techniques have shown significant progress in tackling this problem -- both in the digital \cite{mallappa_rlplace_2022, xu_goodfloorplan_2022, lai_maskplace_2022} and analog \cite{choi_ma-opt_2023, basso_fast_2024, basso_effective_2025} IC design fields.
Although optimal results can be achieved with thorough training, policy fine-tuning is typically required to achieve the best results. This leads to the demand for more data and expensive computational resources.
Methods that do not require pre-training the policy have been suggested in~\cite{zhao_deepth_2023}.
This paper takes a different path to address the problem.
We propose an enhancement of the inference process that builds a search tree to represent the state space of the RL agent from \cite{basso_effective_2025}, then periodically prunes it using the beam search (BS) algorithm~\cite{lowerre_harpy_1976} to balance exploration and exploitation.
This leads to better final results than a 0-shot strategy.
Unlike fine-tuning, it does not require the use of graphics processing units (GPUs) or a policy update, while scaling comparably.
Some previous works used the BS algorithm to solve the bin-packing problem, which is a generalization of block placement in IC floorplanning.
These relied on heuristics and not machine learning \cite{bennell_beam_2010, bennell_beam_2018}.
Similar approaches to ours have been proposed in \cite{choo_simulation-guided_2022, geng_reinforcement_2024}, however, these exploit BS to perform rollouts and policy updates in order to enhance the agent's learning ability. Our solution is plug-and-play and allows us to boost the agent's inference process by leaving untouched its underlying structure. Hybrid BS and RL solutions have been applied to enhance each other's performance \cite{huber_learning_2022, ettrich_policy-based_2023}, but there have not been many applications in Electronic Design Automation (EDA), especially in the case of analog layout.
\begin{figure}
    \centering
    \includegraphics[width=\linewidth]{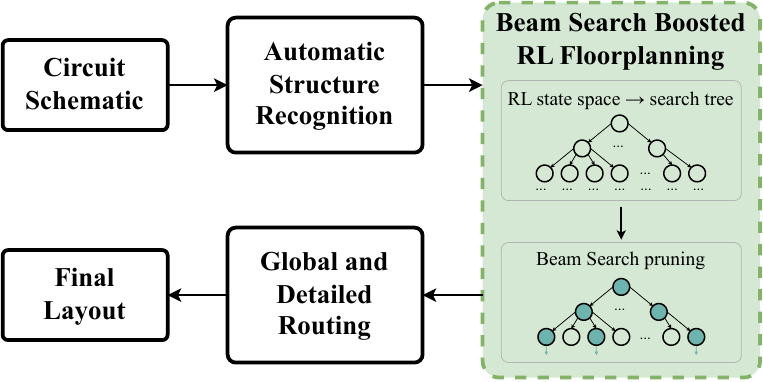}
    \vspace{-0.5cm}
    \caption{High level schematic of the automated layout pipeline presented in \cite{basso_effective_2025}. The contributions of this paper are highlighted in green.}
    \label{flowchart}
\end{figure}
The key contributions of this work are:\looseness=-1
\begin{itemize}
\item A BS enhancement of the RL method proposed in \cite{basso_effective_2025}.
\item A new objective function distinct
from~\cite{basso_effective_2025}.
The users can set the trade-off weights between area and half-perimeter wire length (HPWL) optimization to adjust the output based on their needs.
\item A new congestion management procedure,
inspired by \cite{lai_maskplace_2022},
that ensures a more routing-friendly output and makes the method more suitable for industrial applications.
\item An evaluation of the proposed method in the same testing environment as \cite{basso_effective_2025}. The baseline is improved by 5\% up to 85\% in all metrics comparable to a state-of-the-art (SOtA) fine-tuning approach, and without the use of GPUs.
\end{itemize}
This paper is organized as follows:
The problem is described in Section~\ref{problem_definition},
and our solution is presented in Section~\ref{bs_boosted_floorplanning}.
A comparison is made with previous works in Section~\ref{experimental_results_discussion}.
Conclusions are drawn in Section~\ref{conclusions}.

%% file: paper_body/problem_formulation.tex

As used in~\cite{electronic_design_automation}, let $B\!=\!\{b_1,\ldots,b_i,\ldots, b_m\}$ be a set of rectangular modules with width, height, and area denoted by $w_i$, $h_i$, and $A_i$.
Let $(x_i, y_i)$ denote the bottom-left corner coordinate of module $b_i$ on a chip.
A \emph{floorplan} is an assignment of $(x_i, y_i)$ for each $b_i$, such that no modules overlap.
The goal of the floorplanning problem is to minimize a predefined cost metric.
In our work, this is a weighted normalized sum of several objectives: total area, proxy wire length, and target aspect ratio.
Weights scale the tradeoff between area and other objectives which gives some flexibility to render different outputs as desired by the user.
Routing-friendly floorplans are generated using a procedure (inspired by~\cite{lai_maskplace_2022}) that prevents a preset congestion threshold from being exceeded if possible.

%% file: paper_body/methodology.tex

\subsection{Beam Search}\label{bs}
Beam Search is a heuristic search algorithm that explores a graph in a Breadth-First Search fashion, starting from a source node, and expanding a set composed by the $\beta$ (an integer called \textit{beam width}) most promising successors, for further exploration.
This approach reduces memory and computational requirements, compared to a full search, albeit at the cost of making the search incomplete. Despite this, the algorithm empirically shows good results, providing sub-optimal solutions that are better than the ones obtained with a greedy approach.

\subsection{Reinforcement Learning (RL)}\label{rl}
The RL agent from~\cite{basso_effective_2025} is reused here.
The floorplanning problem is solved by sequential decision-making
using a Markov Decision Process (MDP) denoted by $(S, A_c, P, \mathcal{R}, \gamma)$.
At time step $t$, the agent is in state $s_t \in S$ and chooses an action $a_t\!\in\!A_c$ to execute.
Following the transition probability $p(s_{t+1} | s_t, a_t)\!\in\!P$, the agent moves to a new state $s_{t+1}$ and receives a reward $r_t \in \mathcal{R}$ indicating the impact of its action.
This is discounted by $\gamma$ to balance the relevance of immediate versus future rewards.
The agent eventually learns an optimal policy $\pi^\star(a|s)$ that maximizes the  expected sum of discounted rewards $G_t = \sum_{j=0}^{\infty} \gamma^j\!\cdot\!r_{(t+j+1)}$.
Further details about the encoding of circuit features, training routine, and training data can be found in \cite{basso_effective_2025}. The intermediate reward is defined as:\looseness=-1
\begin{equation}
r_t =
-(\Delta^{DS}_{t} + \Delta^{\text{HPWL}}_{t}),
\end{equation}
$\Delta^{DS}_{t}$ is an increase in relative dead space (DS) between steps:
\begin{equation}
\Delta^{DS}_{t} =
\text{DS}_t\!-\!\text{DS}_{t-1},\ 
\text{DS}_t =
1-\frac{\sum_{i=1}^t A_i}{A^{\mbox{\scriptsize{tot}}}_{t}},
\end{equation}
$A_i$ is the area of module $i$ and $A^{\mbox{\scriptsize{tot}}}_{t}$ is the total floorplan area
in step $t$, $\Delta^{\text{HPWL}}_{t}$ is the increase in HPWL between steps:
\begin{eqnarray}
\Delta^{\text{HPWL}}_{t} \!\!&=&
\text{HPWL}_t-\text{HPWL}_{t-1},\\
\text{HPWL}_t \!\!&=&
\! \sum_{n \in N_t} \scalebox{0.95}{$\max(x_n)-\min(x_n)+\max(y_n)-\min(y_n)$},
\nonumber
\end{eqnarray}
$N_t$ is the subset of nets whose modules have been placed up to step $t$, and $x_n$, $y_n$ are the endpoints of net $n$ in the floorplan.
Given an optional aspect-ratio constraint, $\mbox{Ar}^\star$,
the agent’s end of episode reward is defined as:
\begin{eqnarray}    
r_T = 
-\omega_1 \cdot \frac{1}{1-\text{DS}_T} -\omega_2 \cdot \overline{\text{HPWL}}_T -\omega_3 \cdot \Delta\text{Ar},\\
\overline{\text{HPWL}}_T = \frac{\text{HPWL}_T}{\text{HPWL}_{\min}},\ 
\Delta\text{Ar} = \mbox{Ar}^\star - \mbox{Ar}.\nonumber
\end{eqnarray}
$T$ is the final time step, $\text{HPWL}_{\min}$ is an estimate of the minimum possible \text{HPWL} and is used to normalize the value,
$\Delta\text{Ar}$ is the \emph{outline error} and is the difference between a target aspect ratio $\mbox{Ar}^\star$ and the resulting value $\mbox{Ar}$.
Weights $\omega_1$, $\omega_2$, and $\omega_3$ are empirically set
to balance the impact of area, wire length and aspect ratio on the output floorplan. In this paper, $\omega_1\!=\!1$, $\omega_2\!=\!5$, and $\omega_3\!=\!5$.
Finally, if the generated floorplan violates any predefined constraint, then agent’s behavior is penalized with a reward of $r_t\!=\!-50$.\looseness=-1

\subsection{Beam Search and Reinforcement Learning: BS-RL}\label{bs-rl}
To increase the state space exploration of the 0-shot RL agent, a search tree is built.
The \emph{tree arity $q$} can be chosen by the user to balance exploration. 
Every node of the tree represents a state, and is accompanied by a value $v$, computed as:\looseness=-1
\begin{equation}
v\!=\!\begin{cases}
- \alpha\!\cdot\!\omega_1\!\cdot\!
\displaystyle{\frac{1}{1-\text{DS}_T}}
- \delta\!\cdot\!\omega_2 \cdot \displaystyle{\frac{\text{HPWL}_T}{\text{HPWL}_{\min}}} \ \text{\small for leaves}\\
 -\alpha\!\cdot\!\Delta^{DS}_{t} - \delta\!\cdot\!\Delta^{\text{HPWL}}_{t} \ \ \ \ \ \ \ \ \ \ \ \, \text{\small for internal nodes}
 \end{cases}
\end{equation}
$\alpha$ and $\delta$ are weights that can be set by the user to lead the optimization towards a better area or \text{HPWL}.
This makes the overall procedure more flexible than the original method.
The children of each tree node represent the successor states.
This means that if a node portrays step $t$, then its children describe $q$ possible states at step $t\!+\!1$.
The procedure starts by building the root node $\emptyset$, which represents the initial state $s_0$, then $q$ actions are sampled from $\pi^\star(s_0)$, and the RUDY routing demand estimator is computed, for each sample\cite{spindler_fast_2007}:
\begin{equation}
    \text{RUDY} = \sum_{n \in N_t}d_n \cdot \mathbf{I}_n(x,y),
    \ d_n = \frac{\textit{WA}_n}{\textit{NA}_n},
\end{equation}
$d_n$ is the wire density of net $n$ and is defined as the ratio of wire area, $\textit{WA}_n$, and net area, $\textit{NA}_n$. $\mathbf{I}_n$ is an indicator function over the $x$-$y$-plane for each net, such that\looseness=-1
\begin{equation}
\mathbf{I}_n(x,y)\!=\!\begin{cases}1 \ \text{if}\ 0\!\leq\!x-x_n\!\leq\!w_n \wedge 0\!\leq y-y_n\!\leq\!h_n\\ 0\ \text{otherwise}\ \end{cases},
\end{equation}
where coordinate $(x_n, y_n)$ is the lower left corner of net $n$, while $w_n$ and $h_n$ are the net width and height.
RUDY estimates congestion quickly and accurately and does not depend on a specific routing model.
If this congestion estimate exceeds a threshold set by the user,
then 60\% of the available actions are re-sampled from $\pi^\star(s_0)$, and the ones with highest $\pi^\star(a|s_0)$, that satisfy the congestion threshold, are selected. If no action satisfies this constraint, the ones with the lowest RUDY are used. These actions will generate the children of the root.

After the creation of each level, the tree is pruned with a probability $\epsilon$, that can be chosen by the user to balance exploitation. If this happens, the BS algorithm is executed, leaving (at most) the $\beta$ nodes with the highest $v$ on each level. If not, all the nodes that have been created to this point in the tree, and which have not yet been pruned, are maintained. The tree is then further explored from the remaining nodes in the deepest level created so far. A limit to the width of each level is given as $10^3$ nodes, to reduce the computational and memory burden. If the new level exceeds this threshold, BS is executed to downsize the tree. The building procedure is repeated in the same way for every children node, until the terminal states are found and the leaves are built.\looseness=-1
\begin{figure}
    \centering
    \includegraphics[width=\linewidth]{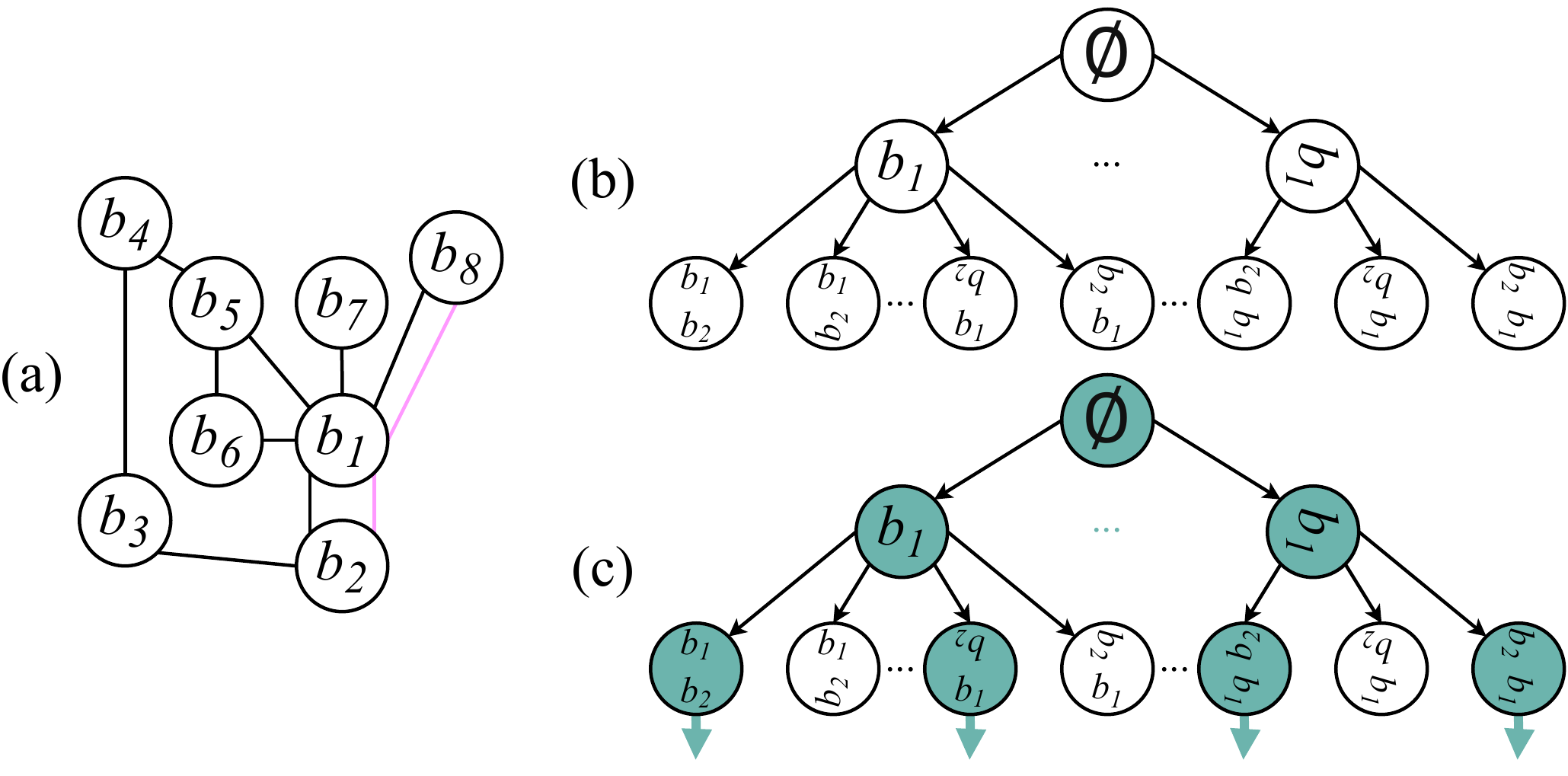}
    \vspace{-0.774cm}
    \caption{\hyperref[fig:bs_example]{
    (a)} OTA-2 graph representation from~\cite[Fig 2]{basso_effective_2025}.
    Pink edges represent alignment constraints. \hyperref[fig:bs_example]{(b)} BS-RL state tree: in step $k\!=\!1$, rectangular module $b_1$ is placed in different ways to form level $1$, $b_2$ is placed in step $k\!=\!2$. \hyperref[fig:bs_example]{(c)} BS is applied after completion of level $2$, with $\beta=4$.}
    \label{fig:bs_example}
\end{figure}
Figure \ref{fig:bs_example} displays a graphic example of the procedure. Once all leaves have been created, the one with the highest $v$ is returned, and the final floorplan computed by BS-RL can be retrieved from its state.

%% file: paper_body/results.tex
\subsection{Experiment Details}
\begin{figure*}
    \centering
    \includegraphics[width=\textwidth]{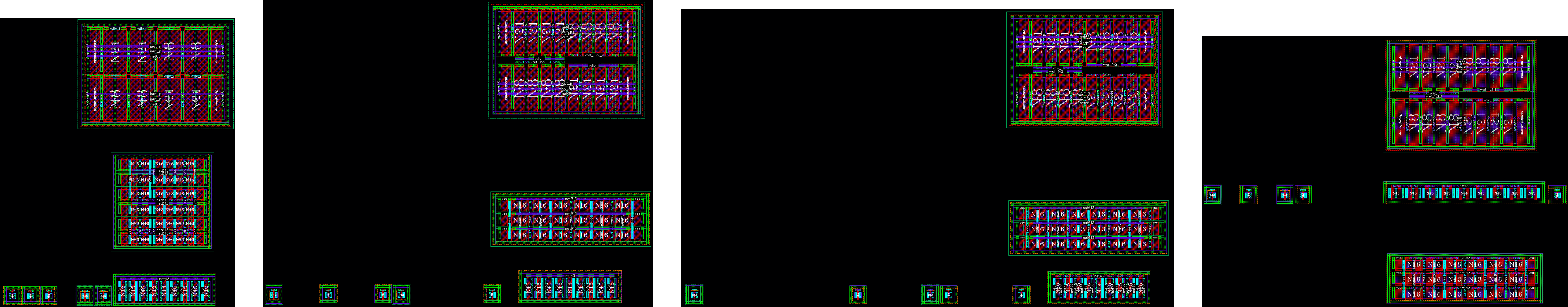}
    \vspace{-0.6cm}
    \caption{Examples of outputs of BS-RL ($k=5$, $\epsilon=0.7$, $\beta=10$), applied to the OTA-2 instance. From left to right, the first two show what happens when setting the congestion threshold to $\infty$ or $0.1$, and the last two illustrate the result of choosing $\alpha=1$ and $\delta=1000$, and vice-versa. Slight manual refinement would be needed before routing.}
    \label{fig:floorplans}
\end{figure*}

Various floorplanning algorithms have been executed on the same IC designs tested in \cite{basso_effective_2025} (Infineon Commercial Technology, $130$nm). In particular, SA is employed as a standard benchmark, the 0-shot RL agent (RL) and the 1000-shot fine-tuned version (RL-FT) from \cite{basso_effective_2025} are used for comparison with BS-RL. Every experiment has been repeated 100 times, computing various metrics (Runtime, DS, HPWL, RUDY, and $\mathcal{R}$). The results have been aggregated with the inter-quartile mean (IQM), to reduce the influence of outliers, and the inter-quartile range (IQR) measures the spread in the results. A confidence interval analysis has been conducted, by computing $10^4$-bootstrap confidence intervals with confidence level $\alpha = 10^{-5}$. Since each statistical measure fell within the corresponding confidence interval, we can confidently infer a high level of statistical significance. BS-RL has been instanced using a tree arity $q = 5$, a pruning probability $\epsilon = 0.7$ and a beam width $\beta = 10$. Although various conﬁgurations were tested, this setup was chosen for evaluation due to its ability to produce good results within a reasonable time frame. Minor adjustments do not signiﬁcantly alter the outcomes, but future investigations may uncover more interesting and potentially superior results. No symmetry, alignment, or aspect-ratio constraints have been specified for experimentation, but Figure \ref{fig:floorplans} shows some output examples, where vertical and horizontal alignment constraints have been specified on the blocks.

Table \ref{tab:results} summarizes the results of these experiments. The grey rows represent IC designs not seen during training by the agent. In BS-RL the congestion threshold is set to $\infty$, to fairly compare it with RL and RL-FT, since they don't implement congestion handling. RL$^\dag$ and RL-FT$^\dag$ overcome this issue, and along BS-RL$^\dag$ they consider a low congestion threshold of $0.1$. The bold values between brackets represent the percentage of improvement from RL($^\dag$) to either RL-FT($^\dag$) or BS-RL($^\dag$). Highlighted values indicate the best results compared to the baseline (\greenhl{with} or \bluehl{without} congestion handling).\looseness=-1

\subsection{Comparison Against Baselines}
\begin{table*}
\centering
\caption{Comparative Analysis of SA, RL Algorithms From \cite{basso_effective_2025}, BS-RL, and Their Variations With Congestion Handling ($^\dag$).}
\vspace{-0.2cm}
\label{tab:results}
\resizebox{\textwidth}{!}{%
\begin{tabular}{@{}ccc!{\color{lightgray}\vrule}c!{\color{lightgray}\vrule}ccc!{\color{lightgray}\vrule}ccc@{}}
\toprule
Circuit & \# Struct. & Metric & SA & RL \cite{basso_effective_2025} & RL-FT \cite{basso_effective_2025} & BS-RL & RL$^\dag$ & RL-FT$^\dag$ & BS-RL$^\dag$ \\ \midrule
 &  & Runtime (s) & \, 0.84 $\pm$ 0.07 & \, \bluehl{0.12 $\pm$ 0.01} & \hspace{-4.75em}147.73 $\pm$ 3.19 & \hspace{-4.25em}17.46 $\pm$ 1.11 & \hspace{-0.05em}\greenhl{0.98 $\pm$ 0.33} & \hspace{-4.38em}149.37 $\pm$ 15.86 & \hspace{-4.25em}18.38 $\pm$ 1.47 \\
 &  & DS (\%) & 48.69 $\pm$ 4.04 & 46.79 $\pm$ 1.70 & \bluehl{43.31 $\pm$ 11.46 \textbf{(3.48\%)}} & \hspace{-0.16em}47.83 $\pm$ 6.35 \textbf{(-1.04\%)} & \hspace{-1.05em}87.12 $\pm$ 6.0 & \hspace{-0.05em}63.79 $\pm$ 18.2 \textbf{(23.33\%)} & \greenhl{48.73 $\pm$ 5.14 \textbf{(38.39\%)}} \\
 &  & HPWL (\textmu m) & 167.94 $\pm$ 25.82 & 224.74 $\pm$ 10.84 & \bluehl{163.91 $\pm$ 43.15 \textbf{(27.07\%)}} & 200.23 $\pm$ 39.81 \textbf{(10.91\%)} & \hspace{0.1em} 560.1 $\pm$ 104.47 & \hspace{0.1em} 284.46 $\pm$ 114.95 \textbf{(49.21\%)} & \greenhl{203.62 $\pm$ 24.02 \textbf{(63.65\%)}} \\
 &  & RUDY & \, \, 3.1 $\pm$ 0.26 & \, 0.19 $\pm$ 0.11 & \, 0.16 $\pm$ 0.05 \textbf{(15.79\%)} & \, \bluehl{0.08 $\pm$ 0.02 \textbf{(57.89\%)}} & \hspace{-0.05em}0.12 $\pm$ 0.01 &  \hspace{0.4em} 0.16 $\pm$ 0.08 \textbf{(-33.33\%)} & \hspace{-0.75em}\greenhl{0.06 $\pm$ 0.01 \textbf{(50\%)}} \\
\multirow{-5}{*}{OTA-1} & \multirow{-5}{*}{5} & $\mathcal{R}$ & \thinspace -2.13 $\pm$ 1.23 & \thinspace -3.88 $\pm$ 0.31 & \thinspace \bluehl{-1.72 $\pm$ 1.78 \textbf{(55.67\%)}} & \thinspace -3.12 $\pm$ 1.21 \textbf{(19.59\%)} & \hspace{-0.875em}-23.44 $\pm$ 2.72 & \hspace{0.05em}-7.02 $\pm$ 5.27 \textbf{(70.05\%)} & \, \thinspace \greenhl{-3.3 $\pm$ 0.88 \textbf{(85.92\%)}} \\ \midrule
 &  & Runtime (s) & \, 0.97 $\pm$ 0.07 & \, \, \bluehl{0.2 $\pm$ 0.02} & \hspace{-4.75em}266.92 $\pm$ 7.32 & \hspace{-4.8em}119.17 $\pm$ 4.87 & \greenhl{3.01 $\pm$ 0.14} & \hspace{-4.375em}274.77 $\pm$ 28.24 & \hspace{-4.25em}251.86 $\pm$ 252.6 \\
 &  & DS (\%) & 54.05 $\pm$ 7.45 & \, 48.31 $\pm$ 13.76 & \hspace{-0.5em}\bluehl{40.03 $\pm$ 4.47 \textbf{(8.28\%)}} & \hspace{-0.53em}46.33 $\pm$ 6.58 \textbf{(1.98\%)} & \hspace{-0.5em}91.64 $\pm$ 1.87 & \hspace{0.4em}66.19 $\pm$ 28.86 \textbf{(25.45\%)} & \, \greenhl{47.67 $\pm$ 10.26 \textbf{(43.97\%)}} \\
 &  & HPWL (\textmu m) & 219.54 $\pm$ 40.71 & 244.04 $\pm$ 54.35 & \bluehl{195.86 $\pm$ 38.07 \textbf{(19.74\%)}} & \hspace{-0.05em}211.89 $\pm$ 37.02 \textbf{(13.17\%)} & \hspace{-0.5em}1001.36 $\pm$ 140.85 & \hspace{0.4em}455.81 $\pm$ 310.63 \textbf{(54.48\%)} & \greenhl{215.86 $\pm$ 77.53 \textbf{(78.44\%)}} \\
&  & RUDY & \, 2.06 $\pm$ 0.56 & \, 0.22 $\pm$ 0.09 & \thinspace \thinspace 0.24 $\pm$ 0.06 \textbf{(-9.09\%)} & \hspace{0.45em}\bluehl{0.18 $\pm$ 0.05 \textbf{(18.18\%)}} & 0.11 $\pm$ 0.01 & \hspace{0.725em}0.14 $\pm$ 0.08 \textbf{(-27.27\%)} & \, \greenhl{0.1 $\pm$ 0.03 \textbf{(9.09\%)}} \\
\multirow{-5}{*}{OTA-2} & \multirow{-5}{*}{8} & $\mathcal{R}$ & \thinspace -3.02 $\pm$ 1.45 & \thinspace -4.12 $\pm$ 2.32 & \thinspace \bluehl{-2.19 $\pm$ 1.14 \textbf{(46.84\%)}} & \hspace{-0.4em}-2.95 $\pm$ 1.41 \textbf{(28.4\%)} & \hspace{-0.825em}-39.86 $\pm$ 6.17 & \hspace{0.05em}-12.43 $\pm$ 13.29 \textbf{(68.82\%)} & \thinspace \greenhl{-3.17 $\pm$ 3.02 \textbf{(92.05\%)}} \\ \midrule
 &  & Runtime (s) & \, 1.24 $\pm$ 0.07 & \, \bluehl{0.23 $\pm$ 0.02} & \hspace{-4.3em}282.63 $\pm$ 30.86 & \hspace{-4.3em}76.69 $\pm$ 2.94 & \greenhl{4.76 $\pm$ 0.38} & \hspace{-4.35em}286.93 $\pm$ 31.48 & \hspace{-3.75em}131.95 $\pm$ 160.59 \\
 &  & DS (\%) & 63.95 $\pm$ 6.32 & 51.26 $\pm$ 6.14 & \thinspace \thinspace 55.54 $\pm$ 10.65 \textbf{(-4.28\%)} & \hspace{-1.75em}\bluehl{48.26 $\pm$ 9.49 \textbf{(3\%)}} & \hspace{-0.525em}91.73 $\pm$ 0.52 & \hspace{0.45em}76.85 $\pm$ 15.02 \textbf{(14.88\%)} & \greenhl{51.16 $\pm$ 7.52 \textbf{(40.57\%)}} \\
 &  & HPWL (\textmu m) & 286.92 $\pm$ 53.53 & 337.36 $\pm$ 65.07 & \bluehl{266.79 $\pm$ 79.72 \textbf{(20.92\%)}} & \hspace{-0.5em}270.66 $\pm$ 43.2 \textbf{(19.77\%)} & \hspace{-0.525em}1016.39 $\pm$ 142.79 & \hspace{0.45em}446.54 $\pm$ 164.21 \textbf{(56.07\%)} & \greenhl{284.25 $\pm$ 62.75 \textbf{(72.03\%)}} \\
&  & RUDY & \, 5.01 $\pm$ 0.54 & \, 0.37 $\pm$ 0.24 & \hspace{-1.75em}0.37 $\pm$ 0.1 \textbf{(0\%)} & \thinspace \thinspace \thinspace \bluehl{0.16 $\pm$ 0.04 \textbf{(56.76\%)}} & \hspace{-0.05em}0.15 $\pm$ 0.04 & \, 0.3 $\pm$ 0.14 \textbf{(-100\%)} & \hspace{-0.25em}\greenhl{0.1 $\pm$ 0 \textbf{(33.33\%)}} \\
\multirow{-5}{*}{Bias-1} & \multirow{-5}{*}{9} & $\mathcal{R}$ & \thinspace -6.48 $\pm$ 2.28 & \, \thinspace -8.2 $\pm$ 2.51 & \thinspace -5.84 $\pm$ 3.42 \textbf{(28.78\%)} & \thinspace \bluehl{-5.72 $\pm$ 1.52 \textbf{(30.24\%)}} & \hspace{-0.4em}-42.4 $\pm$ 4.99 & \hspace{-0.4em}-14.52 $\pm$ 8.68 \textbf{(65.75\%)} & \thinspace \greenhl{-6.25 $\pm$ 2.03 \textbf{(85.26\%)}} \\ \midrule
\rowcolor[HTML]{EFEFEF} 
\cellcolor[HTML]{EFEFEF} & \cellcolor[HTML]{EFEFEF} & Runtime (s) & \, 0.88 $\pm$ 0.05 & \, \bluehl{0.17 $\pm$ 0.02} & \hspace{-4.25em}200.84 $\pm$ 23.22 & \hspace{-4.75em}216.59 $\pm$ 7.03 & \greenhl{2.74 $\pm$ 0.37} & \hspace{-4.35em}201.34 $\pm$ 23.42 & \hspace{-4.25em}357.69 $\pm$ 43.79 \\
\rowcolor[HTML]{EFEFEF} 
\cellcolor[HTML]{EFEFEF} & \cellcolor[HTML]{EFEFEF} & DS (\%) & 62.79 $\pm$ 6.21 & 52.22 $\pm$ 8.12 & 49.81 $\pm$ 12.83 \textbf{(2.41\%)} & \hspace{-0.5em}\bluehl{45.61 $\pm$ 8.58 \textbf{(6.61\%)}} & \hspace{-1em}90.67 $\pm$ 2.2 & \hspace{0.4em}76.13 $\pm$ 14.19 \textbf{(14.54\%)} & \, \, \greenhl{50.9 $\pm$ 10.58 \textbf{(39.77\%)}} \\
\rowcolor[HTML]{EFEFEF} 
\cellcolor[HTML]{EFEFEF} & \cellcolor[HTML]{EFEFEF} & HPWL (\textmu m) & \, 124.2 $\pm$ 22.31 & 128.44 $\pm$ 15.78 & \hspace{-1em}123.83 $\pm$ 25.2 \textbf{(3.59\%)} & \hspace{-0.5em}\bluehl{121.21 $\pm$ 10.25 \textbf{(5.63\%)}} & \hspace{-0.5em}336.37 $\pm$ 68.91 & \hspace{-0.075em}204.05 $\pm$ 69.18 \textbf{(39.34\%)} & \greenhl{130.57 $\pm$ 16.37 \textbf{(61.18\%)}} \\
\rowcolor[HTML]{EFEFEF}
\cellcolor[HTML]{EFEFEF} & \cellcolor[HTML]{EFEFEF} & RUDY & \, 3.76 $\pm$ 1.79 & \, 0.68 $\pm$ 0.51  & \, \thinspace \thinspace 0.94 $\pm$ 0.45 \textbf{(-38.24\%)} & \hspace{0.5em}\bluehl{0.26 $\pm$ 0.12 \textbf{(61.76\%)}} & \greenhl{0.12 $\pm$ 0.02} & \, 0.6 $\pm$ 0.36 \textbf{(-400\%)} & \hspace{0.85em}0.16 $\pm$ 0.07 \textbf{(-33.33\%)} \\
\rowcolor[HTML]{EFEFEF}
\multirow{-5}{*}{\cellcolor[HTML]{EFEFEF}RS Latch} & \multirow{-5}{*}{\cellcolor[HTML]{EFEFEF}7} & $\mathcal{R}$ & \, \thinspace -5.1 $\pm$ 2.25 & \thinspace -4.96 $\pm$ 1.54 & \hspace{-0.82em}-4.63 $\pm$ 2.2 \textbf{(6.65\%)} & \hspace{0.16em}\bluehl{-4.19 $\pm$ 0.92 \textbf{(15.52\%)}} & \hspace{-0.8em}-28.25 $\pm$ 6.65 & \hspace{-0.375em}-12.54 $\pm$ 7.52 \textbf{(55.61\%)} & \thinspace \greenhl{-4.96 $\pm$ 1.57 \textbf{(82.44\%)}} \\ \midrule
\rowcolor[HTML]{EFEFEF} 
\cellcolor[HTML]{EFEFEF} & \cellcolor[HTML]{EFEFEF} & Runtime (s) & \, 1.61 $\pm$ 0.06 & \, \bluehl{0.45 $\pm$ 0.04} & \hspace{-4.25em}696.88 $\pm$ 71.18 & \hspace{-4.3em}254.64 $\pm$ 10.68 & \hspace{-0.5em}\greenhl{15.04 $\pm$ 1.43} & \hspace{-4.3em}848.91 $\pm$ 68.34 & \hspace{-3.75em}395.39 $\pm$ 279.74 \\
\rowcolor[HTML]{EFEFEF} 
\cellcolor[HTML]{EFEFEF} & \cellcolor[HTML]{EFEFEF} & DS (\%) & 71.75 $\pm$ 9.59 & 72.62 $\pm$ 8.66 & \, \bluehl{54.34 $\pm$ 11.55 \textbf{(18.28\%)}} & 57.56 $\pm$ 8.63 \textbf{(15.06\%)} & 91.8 $\pm$ 0.77 & \, 61.62 $\pm$ 11.84 \textbf{(30.18\%)} & \greenhl{57.78 $\pm$ 7.01 \textbf{(34.02\%)}} \\
\rowcolor[HTML]{EFEFEF} 
\cellcolor[HTML]{EFEFEF} & \cellcolor[HTML]{EFEFEF} & HPWL (\textmu m) & 1931.84 $\pm$ 407.06 & \, 2502.8 $\pm$ 374.68 & \bluehl{1647.64 $\pm$ 290.09 \textbf{(34.17\%)}} & 1767.78 $\pm$ 329.69 \textbf{(29.37\%)} & \hspace{-0.5em}5849.24 $\pm$ 878.71 & 1878.77 $\pm$ 369.52 \textbf{(67.88\%)} & \greenhl{1791.86 $\pm$ 279.63 \textbf{(69.37\%)}} \\
\rowcolor[HTML]{EFEFEF} 
\cellcolor[HTML]{EFEFEF} & \cellcolor[HTML]{EFEFEF} & RUDY & \, 4.66 $\pm$ 0.88 & \, 0.25 $\pm$ 0.13 & \hspace{-0.44em}0.32 $\pm$ 0.24 \textbf{(-28\%)} & \hspace{-0.73em}\bluehl{0.08 $\pm$ 0.01 \textbf{(68\%)}} & 0.27 $\pm$ 0.09 & \hspace{0.8em}0.32 $\pm$ 0.25 \textbf{(-18.52\%)} & \hspace{0.5em}\greenhl{0.07 $\pm$ 0.01 \textbf{(74.07\%)}} \\
\rowcolor[HTML]{EFEFEF}
\multirow{-5}{*}{\cellcolor[HTML]{EFEFEF}Driver} & \multirow{-5}{*}{\cellcolor[HTML]{EFEFEF}17} & $\mathcal{R}$ & \thinspace -8.35 $\pm$ 3.22 & \thinspace -13.1 $\pm$ 2.82 & \thinspace \bluehl{-6.26 $\pm$ 2.18 \textbf{(52.21\%)}} & \thinspace -7.16 $\pm$ 2.48 \textbf{(45.34\%)} & \hspace{-0.825em}-42.18 $\pm$ 6.19 & \hspace{0.16em}-8.16 $\pm$ 2.74 \textbf{(80.65\%)} & \hspace{0.15em}\greenhl{-7.26 $\pm$ 2.22 \textbf{(82.79\%)}} \\ \midrule
\rowcolor[HTML]{EFEFEF} 
\cellcolor[HTML]{EFEFEF} & \cellcolor[HTML]{EFEFEF} & Runtime (s) & 1.91 $\pm$ 0.1 & \, \bluehl{0.52 $\pm$ 0.05} & \hspace{-4.25em}983.98 $\pm$ 94.55 & \hspace{-4.75em}1293.42 $\pm$ 82.25 & \hspace{-0.5em}\greenhl{13.53 $\pm$ 0.98} & \hspace{-4.775em}1182.4 $\pm$ 62.7 & \hspace{-4.25em}1490.64 $\pm$ 321.83 \\
\rowcolor[HTML]{EFEFEF} 
\cellcolor[HTML]{EFEFEF} & \cellcolor[HTML]{EFEFEF} & DS (\%) & 66.05 $\pm$ 5.92 & 75.73 $\pm$ 5.39 & \, \bluehl{62.1 $\pm$ 7.89 \textbf{(13.63\%)}} & \hspace{-0.5em}71.55 $\pm$ 6.11 \textbf{(4.18\%)} & \hspace{-0.5em}91.12 $\pm$ 0.73 & \, \greenhl{63.75 $\pm$ 10.09 \textbf{(27.37\%)}} & 71.56 $\pm$ 6.14 \textbf{(19.56\%)} \\
\rowcolor[HTML]{EFEFEF} 
\cellcolor[HTML]{EFEFEF} & \cellcolor[HTML]{EFEFEF} & HPWL (\textmu m) & 2463.06 $\pm$ 571.35 & 3576.77 $\pm$ 520.26 & \bluehl{2694.18 $\pm$ 429.98 \textbf{(24.68\%)}} & 2953.08 $\pm$ 423.86 \textbf{(17.44\%)} & 8532.99 $\pm$ 1306.63 & \greenhl{2815.12 $\pm$ 550.38 \textbf{(67.01\%)}} & 2921.03 $\pm$ 468.06 \textbf{(65.77\%)} \\
\rowcolor[HTML]{EFEFEF} 
\cellcolor[HTML]{EFEFEF} & \cellcolor[HTML]{EFEFEF} & RUDY & \, 2.33 $\pm$ 0.39 & \, 0.12 $\pm$ 0.05 & \hspace{-0.43em}0.15 $\pm$ 0.08 \textbf{(-25\%)} & \hspace{-0.75em}\bluehl{0.06 $\pm$ 0.01 \textbf{(50\%)}} & 0.24 $\pm$ 0.11 & \hspace{0.5em}0.11 $\pm$ 0.04 \textbf{(54.17\%)} & \hspace{-0.75em}\greenhl{0.06 $\pm$ 0.01 \textbf{(75\%)}} \\
\rowcolor[HTML]{EFEFEF}
\multirow{-5}{*}{\cellcolor[HTML]{EFEFEF}Bias-2} & \multirow{-5}{*}{\cellcolor[HTML]{EFEFEF}19} & $\mathcal{R}$ & \thinspace -2.99 $\pm$ 2.01 & \thinspace -8.12 $\pm$ 1.88 & \hspace{0.155em}\bluehl{-4.07 $\pm$ 1.56 \textbf{(49.88\%)}} & \hspace{0.16em}-5.58 $\pm$ 1.37 \textbf{(31.28\%)} & \hspace{-0.825em}-28.73 $\pm$ 3.43 & \hspace{0.18em}\greenhl{-4.64 $\pm$ 1.64 \textbf{(83.85\%)}} & \thinspace -5.55 $\pm$ 1.37 \textbf{(80.68\%)} \\ \bottomrule
\end{tabular}%
}
\vspace{-0.25cm}
\end{table*}

As expected, SA produces very compact floorplans, but their very high RUDY estimate suggests low practical usability, and the need for a lot of manual refinement. RL, and in particular RL-FT, generate very compact but less congested floorplans.
The improvement recorded by BS-RL aligns with the one given by RL-FT, sometimes even surpassing it. Moreover, the runtime of BS-RL is generally similar to RL-FT, often lower, but it has the advantage of running on CPU, while RL-FT needs GPUs to compute in a reasonable time. The only outlier is observed for the Bias-2 instance, but as suggested by the runtimes of smaller problem instances, it doesn't scale monotonically with respect to problem size. Supposedly, runtime doesn't increase significantly on problems of even higher complexity, but new data would be needed for assessment. Some negative percentage improvements are recorded, in particular for the RUDY in the RL-FT and RL-FT$^\dag$ columns. Since fine-tuning produces more compact outputs, it's likely that these floorplans don't leave much space for routing. Also the BS-RL$^\dag$ result of the RS Latch instance records a negative improvement, but the absolute value is close to the one accomplished by RL$^\dag$, meaning that optimality is probably achieved by both procedures. One of the most surprising results of this experimental evaluation is that the outputs of RL$^\dag$, compared to RL, even though are better in terms of RUDY, are far worse in terms of DS, HPWL and $\mathcal{R}$. With BS it is possible to obtain floorplans that are less congested, and as compact as the ones obtained without requiring the congestion constraint.\looseness=-1

To conclude, the BS-RL($^\dag$) algorithm improves the results obtained with RL($^\dag$), having average $\%$ improvement across all test cases: DS: $4.97\%$, HPWL: $16.05\%$, RUDY: $52.09\%$, $\mathcal{R}$: $28.39\%$ (DS: $36.05\%$, HPWL: $68.41\%$, RUDY: $34.69\%$, $\mathcal{R}$: $84.86\%$). Moreover, the measured metrics align closely with the ones achieved by a SOtA procedure like RL-FT($^\dag$), and often show higher stability, having lower IQR.

%% file: paper_body/conclusions_future_research.tex

This paper presents an enhancement to the methodology of \cite{basso_effective_2025}. By exploiting a BS strategy, exploration of the RL agent is increased in a simple way, without changing the underlying structure, to the point that it becomes comparable with a fine-tuned agent. The original solution's generalization ability, handling of circuit features, overlap avoidance, and management of various constraints stay intact. This floorplanning algorithm is also capable of leading optimization towards different objectives, as wished by the user. A congestion management procedure renders more routing friendly outputs. These improvements do not necessitate updating or fine-tuning the policy, thus eliminating the need for GPUs, and they do not compromise runtime efficiency.

In the future, taking inspiration from what human engineers do in analog layout, we aim to augment the floorplan algorithm with detailed routing information, and feedbacks from post-layout verification, to incrementally refine the device placement by re-iterating the automated layout pipeline. The goal is, at some point, to achieve automatically generated LVS, DRC and ERC clean analog layout designs.